\documentclass[conference]{IEEEtran}
\IEEEoverridecommandlockouts
\usepackage{cite}
\usepackage{amsmath,amssymb,amsfonts}
\usepackage{algorithmic}
\usepackage{graphicx}
\usepackage{textcomp}
\usepackage{xcolor}
\usepackage{libertine}
\usepackage{tikz}
\usepackage[super]{nth}
\usepackage{multirow} 
\usepackage[most]{tcolorbox}
\def\BibTeX{{\rm B\kern-.05em{\sc i\kern-.025em b}\kern-.08em
    T\kern-.1667em\lower.7ex\hbox{E}\kern-.125emX}}

\definecolor{darkgreen}{RGB}{0,120,0}

\DeclareMathOperator*{\argmin}{arg\,min}
    
\begin{document}

\title{\texttt{SONATA}: \underline{S}elf-adaptive Ev\underline{o}lutionary Framework for Hardware-aware \underline{N}eural \underline{A}rchitec\underline{t}ure Se\underline{a}rch}

\author{
\IEEEauthorblockN{
Halima Bouzidi\IEEEauthorrefmark{1},
Smail Niar\IEEEauthorrefmark{1},
Hamza Ouarnoughi\IEEEauthorrefmark{1},
El-Ghazali Talbi\IEEEauthorrefmark{2}}

\IEEEauthorrefmark{1}\textit{LAMIH/UMR CNRS, Université Polytechnique Hauts-de-France, Valenciennes, France} \\
\IEEEauthorrefmark{2}\textit{University of Lille, Centre de Recherche en Informatique Signal et Automatique de Lille (CRIStAL)} \\

\IEEEauthorrefmark{1}\textit{\{firstname.lastname\}}@uphf.fr
\hspace{20truemm}
\IEEEauthorrefmark{2}\textit{\{firstname.lastname\}}@univ-lille.fr
\vspace{-5truemm}
}

\maketitle

\begin{abstract}
Recent advancements in Artificial Intelligence (AI), driven by Neural Networks (NN), demand innovative neural architecture designs, particularly within the constrained environments of Internet of Things (IoT) systems, to balance performance and efficiency. HW-aware Neural Architecture Search (HW-aware NAS) emerges as an attractive strategy to automate the design of NN using multi-objective optimization approaches, such as evolutionary algorithms. However, the intricate relationship between NN design parameters and HW-aware NAS optimization objectives remains an underexplored research area, overlooking opportunities to effectively leverage this knowledge to guide the search process accordingly. Furthermore, the large amount of evaluation data produced during the search holds untapped potential for refining the optimization strategy and improving the approximation of the Pareto front. Addressing these issues, we propose SONATA, a self-adaptive evolutionary algorithm for HW-aware NAS. Our method leverages adaptive evolutionary operators guided by the learned importance of NN design parameters. Specifically, through tree-based surrogate models and a Reinforcement Learning agent, we aspire to gather knowledge on '\textit{How}' and '\textit{When}' to evolve NN architectures. Comprehensive evaluations across various NAS search spaces and hardware devices on the ImageNet-1k dataset have shown the merit of SONATA with up to \textbf{$\sim$0.25\%} improvement in accuracy and up to \textbf{$\sim$2.42x} gains in latency and energy. Our SONATA has seen up to \textbf{$\sim$93.6\%} Pareto dominance over the native NSGA-II, further stipulating the importance of self-adaptive evolution operators in HW-aware NAS.
\end{abstract}

\vspace{0.2cm}


\section{Introduction}
Neural Networks (NNs) have catalyzed the AI innovation we are witnessing today in myriad application fields such as health care, smart cities, and transportation. The advent of Edge computing systems has further expanded the deployment of NNs across a broad spectrum of Internet of Things (IoT) devices, ranging from high-performance Edge GPUs to Tiny micro-controllers. In front of this diverse wave of tasks, applications, systems, and computing devices, the architecture design of NNs is pivotal for achieving optimal performance and efficiency. Formally, the design process of NNs can be articulated as a bi-level optimization problem where neural architecture design parameters (i.e., layers and operators) and neuron weights are both searchable. However, tuning the NN design parameters, learning the neurons' weights, and choosing the adequate hardware configuration for deployment is labor intensive. Furthermore, neural design domains lack interpretability and explainability regarding the obtained performances \cite{yu2019evaluating}. Recently, efforts have been shifted towards automating the design process of NN with Hardware-awareness via the '\textit{HW-aware NAS}' paradigm \cite{cai2018proxylessnas, arxivHadjer}. Still, the search space of NNs is typically fractal, relatively large, and difficult to explore because of the time-consuming trial-error iterative evaluation process that requires an expensive optimization budget (i.e., search time and computing resources) to measure performance on the target task/dataset and efficiency related-metric on the target hardware device.

\begin{figure}[htp]
\centering
    \includegraphics[width=.48\textwidth]{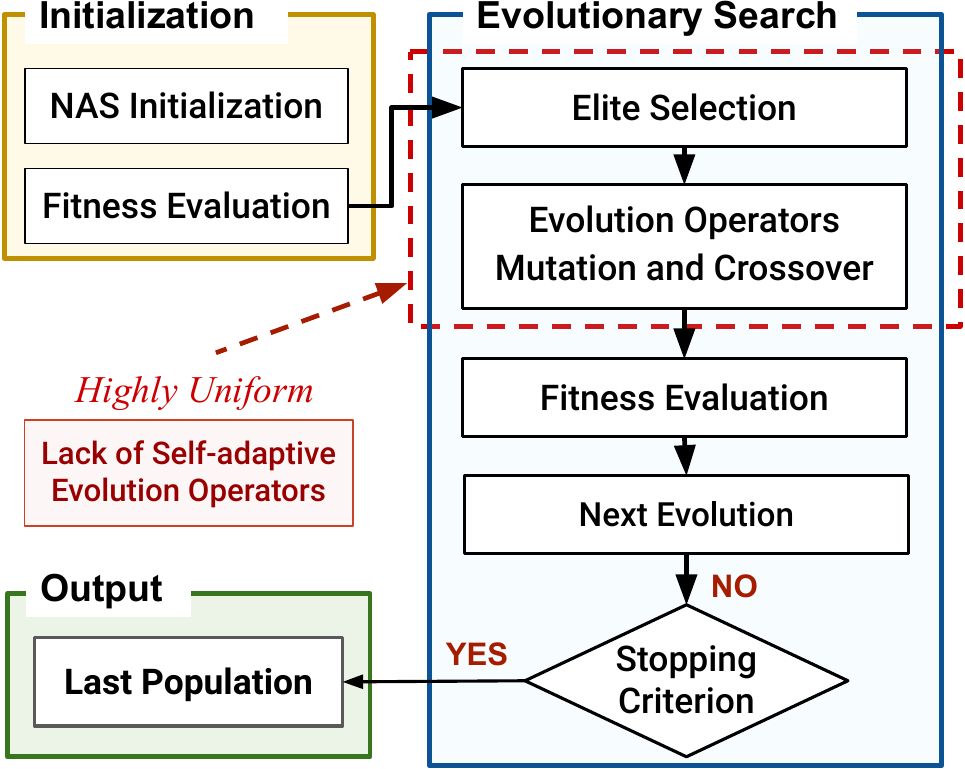}
    \caption{The workflow of a typical Evolutionary NAS framework (ENAS). 
    }
    \label{fig:sonata_motiv}
\end{figure}

Earlier \textit{HW-aware NAS} approaches predominantly utilized random search to circumvent the stagnation in local optima \cite{chen2018searching}. Nevertheless, these approaches were deemed ineffective and inefficient as NN architectures are generated randomly and explored unthinkingly without a well-established basis. Subsequently, evolutionary search approaches have been adopted to lower the uncertainty by contributing fitness evaluation functions that guide the search process to favor exploiting of the best-performing NNs and exploring randomly generated ones. Following this mechanism, NN neighborhoods are generated in a manner that jeopardizes both exploration and exploitation \cite{lu2020nsganetv2}. Despite these advancements, the efficacy of such evolutionary search strategies remains constrained by their over-reliance on highly uniform evolutionary operators such as '\textit{mutation}' and '\textit{crossover}' to produce new populations of NNs (See Figure \ref{fig:sonata_motiv}). Within this conventional framework, mutation is randomly applied to \textit{add}, \textit{alter}, or \textit{remove} neural components to create a new \textit{offspring} population. Similarly, in crossover, neural components are randomly selected and recombined from a set of predefined optimal NN architectures to produce new ones.

Both \textit{mutation} and \textit{crossover} in evolutionary search are founded upon the basis of '\textit{natural selection}' -- assuming that \textit{the mutation or recombination of genes from well-designed genomes enhances the design and performances of the resulting child genome}. However, this assumption does not always hold regarding NP-hard problems, such as \textit{HW-aware NAS}, as uniform evolutionary operators between well-performing NN architectures may result in poorly performing ones. Consequently, this can steer the search process towards ill-designed NNs with worse performances, wasting an expensive optimization budget. Thus, carefully choosing the criteria of '{\it{Where and How}}' to apply the evolutionary operators is necessary for strategic optimization. Such expertise can be acquired in a data-driven fashion during the search by exploiting the history of evaluated past populations to enhance the knowledge of the NN search space and better guide the evolutionary algorithm toward global optima. 

\begin{tcolorbox}[colback=blue!2!white,colframe=white!7!white]
\textit{Evolutionary algorithms must operate through incremental, strategical adaptations, leveraging pre-existing NN architectures rather than embodying the role of an expert designer crafting optimal NN designs de novo.}
\end{tcolorbox}

\subsection{Identified Research Problems}
Data-driven modeling approaches are widely used to improve the design of NNs following the \textit{ML-for-ML} paradigm. For instance, data-driven surrogate models leverage evaluation data of NNs to train a model capable of estimating the performances of unseen ones \cite{liu2022survey}. Nonetheless, this approach has primarily been limited to NN performance estimation to accelerate the fitness evaluation in evolutionary HW-aware NAS \cite{lu2020nsganetv2, greenwood2022surrogate}. Viewing things from a fresh angle -- data-driven surrogate models can also be used to learn and infer the importance of NN design parameters to guide the search operators in Evolutionary NAS (ENAS) frameworks. Specifically, importance scores can be used to determine which design parameters to evolve to maximize the gains in performance and efficiency. Incorporating such information into existing HW-aware ENAS frameworks could further demystify their \textit{black-box} nature and shift the paradigm towards one that provides a higher level of explainability regarding the quality of newly designed NN architectures.

As far as we know, the systematic utilization of \textit{the history of evaluation data} from previously explored NN architectures during the evolutionary search has not been thoroughly studied in the literature -- especially in the multi-objective context of HW-aware NAS. The history of evaluated NN populations may help build and update data-driven ML-based surrogate models capable of guiding the search algorithm and assessing the importance of NN design parameters. However, overheads must be minimized to employ this approach efficiently and maximize its benefits. For this reason, the additional execution time, energy consumption, and memory usage for data storage and surrogate model training should be carefully considered for such surrogate-assisted approaches.

\subsection{Novel Scientific Contributions}
In the realm of our observations and motivations, we introduce the following novel scientific contributions:
\begin{itemize}
    \item We present \texttt{SONATA}, a self-adaptive multi-objective optimization framework for HW-aware-NAS that progressively learns the importance of NN design parameters and guides the search strategy accordingly.
    \item We design and implement self-adaptive evolutionary search operators guided by ML-based models to continuously learn the importance of NN design parameters from the search data history such that to explore and generate effective neighborhoods of NNs by mutating and crossover the most critical neural design parameters.
    \item We compare \texttt{SONATA} against baseline evolutionary optimization algorithms for HW-aware-NAS frameworks. 
    \item We validate our approach on many HW-aware NAS problem instances by varying the search space and the target hardware device for the image classification task on the large ImageNet-1k dataset.
    \item Through extensive experiments, we demonstrate that our approach can find optimal NNs under a low optimization budget. Evaluation results have shown the merit of \texttt{SONATA} with an accuracy improvement up to \textbf{$\sim$0.25\%} on the ImageNet-1k dataset and latency/energy gains up to \textbf{$\sim$2.42x} on Edge GPUs. Under the same optimization budget (i.e. number of evolutionary iterations), \texttt{SONATA} has also seen up to \textbf{$\sim$93.6\%} Pareto dominance over the native SOTA algorithm, NSGA-II.
\end{itemize}

\section{Related Work}\label{sec:related_works}

\subsection{Neural Architecture Search (NAS)}
NAS aims to automate the design of NN by performing an iterative search to find a suitable NN architecture for a target task \cite{elsken2019neural}. Typical NAS frameworks can be framed as a black-box optimization that takes as input a predefined search space, search strategy, and fitness evaluation function. These three components make each NAS framework unique in its functionality and directly impact the quality of its output results. To enhance the performance of NAS, many contributions have been made in the literature to improve the three building blocks mentioned above. The search space is one critical component as it defines the scope of the \textit{to-explore} NN designs \cite{yu2019evaluating}. Search spaces can be classified according to the granularity of the searchable parameter to: 
\begin{enumerate}
    \item \textit{Macro-search space}, in which the NN structure and operators are fully searchable \cite{ying2019bench, dong2019bench}.
    \item \textit{Micro-search space}, in which the NN structure is fixed, and operators are searchable \cite{cai2019once, wu2019fbnet}.
\end{enumerate}

The next important NAS component is the search strategy. Evolutionary algorithms \cite{lorenzo2017particle, lu2019nsga} and Reinforcement Learning \cite{hsu2018monas} approaches are widely employed in existing NAS works. Bayesian optimization has also shown impressive results \cite{white2021bananas, shen2023proxybo} but needs to be adapted to the discrete and categorical nature of NAS encoding schemes as its surrogates and acquisition functions are limited to continuous domains. 

Despite careful choices of search space and strategy, the fitness evaluation in NAS is the real bottleneck. It typically consumes more than 95\% of the search overhead, especially when hardware deployment is considered in the loop. Many efforts have been devoted to accelerating this part by developing proxy estimators such as performance predictors \cite{siems2020bench, white2021powerful, bouzidi2022performance}, look-up-tables \cite{li2020hw}, multi-fidelity estimators and knowledge distillation \cite{chen2022mfenas, trofimov2023multi}, network morphism \cite{cai2019once, cai2018efficient} one-shot NAS \cite{liu2018darts, wang2021idarts, pham2018efficient}, and zero-shot NAS \cite{lin2021zen, yan2021zeronas}.

\subsection{Evolutionary Neural Architecture Search (ENAS)}
Evolutionary algorithms are employed in many NAS frameworks due to their efficiency and flexibility in handling large search spaces with complex encoding schemes \cite{liu2021survey}. ENAS is a derivative-free optimization algorithm that can converge faster in the early stages and perform as well as other algorithms in later stages. As typical evolutionary search methods, ENAS requires extensive evaluation and trial-error iterations, which turns out to be time-consuming for HW-aware NAS problems \cite{pan2021neural}. Hence, employing performance predictors in the ENAS framework -- in a surrogate-assisted manner -- can reduce the search time \cite{pan2021neural}. The key components of ENAS involve the following: 
\begin{itemize}
    \item (\emph{i}) Encoding: Define how a NN architecture is represented. ENAS generally operates on the adjacency matrix of the NN using a one-hot/categorical encoding \cite{white2020study} or by learning an embedding of the directed acyclic graph representation using GCN or GNN  \cite{shi2020bridging}.
    \item (\emph{ii}) Mutation: Dictate the set of design modifications to be made on a given NN architecture to generate a new one through adding, altering, or removing neural components (e.g., layers, operators, inter-layer connection).
    \item (\emph{iii}) Crossover: Determine how two NN architectures are recombined to derive a new one.
    \item (\emph{iv}) Elite selection: To render a set of Pareto optimal NN architectures as elite solutions that will undergo \textit{mutation} and \textit{crossover} steps \cite{sinha2022novelty}. ENAS generally employ Pareto ranking methods to sort solutions according to their dominance and diversity in the multi-objective space using tournament selection and crowding distance measurements \cite{deb2000fast}.
\end{itemize} 

\subsection{Surrogate-assisted Multi-objective ENAS (SaMo-ENAS)}
Given the high convexity observed in HW-aware NAS problems, simple analytical models can not be used to infer insights on the causality and correlation between NN architectures and the obtained performances. Therefore, ML-based surrogate models\footnote{'Surrogate models', 'Prediction models', or 'Predictors' can be used interchangeably to designate performance estimation models. The terminology of '\textit{Surrogate models}' is widely used in the context of HW-aware NAS.} have been leveraged to accelerate the ENAS process \cite{fan2023surrogate}. These surrogate models can be used at different levels of the ENAS algorithm, including: 
\begin{itemize}
    \item (\emph{i}) \textit{Fitness evaluation}: Using performance predictors during the search to estimate the accuracy, latency, or energy of the explored NNs. The training process of the performance predictors can be held in an:
    \begin{itemize}
        \item Offline setting (i.e., Prior to the search),
        \item Online setting (i.e., During the search) 
    \end{itemize}
        
  In the literature, surrogate models are typically built upon Random forest \cite{peng2022pre}, XGBoost \cite{pan2021neural}, GNN \cite{wei2022npenas}, Gaussian process \cite{garrido2020dealing, calisto2021emonas, cho2022b2ea}, or ensemble methods \cite{wu2021stronger}.
    \item (\emph{ii}) \textit{Initialization and sampling}: By gathering pre-knowledge on the quality of the search space to enhance the sampling of NNs using multi-fidelity estimators \cite{wang2021sample}, Gaussian process \cite{cho2022b2ea}, or clustering methods \cite{traore2023data}.
    \item (\emph{iii}) Neighborhood generation: In ENAS, neighborhoods are generated to produce the next population of NNs by balancing the exploration-exploitation tradeoff through mutation and crossover. In that sense, \cite{chen2019renas} was the first paper to introduce a learning-based mutation using a Reinforcement Learning agent. Probabilistic models of mutation have been introduced in \cite{xue2021self} to self-adapt the mutation strategy during the search in a neural block-wise level of the NN. In that same spirit, \cite{qiu2023efficient} proposes a mutation  agent based on the NN model size. 
\end{itemize}

However, most existing works still need to explore the prospect of data-driven methods to learn the importance of NN design parameters for HW-aware NAS. Such knowledge and expertise are highly required, mainly since not all neural design parameters contribute equally to the contradictory optimization objectives of HW-aware NAS. By pinpointing the most important design parameters, one can use this information to shrink/enlarge the search space or strategically guide the search algorithm. In this paper, we mainly focus on the second perspective. To the best of our knowledge, our work is the first to address this issue for HW-aware evolutionary NAS by proposing a purely data-driven approach that exploits the search history in ENAS. At design time and in parallel with the search process, a surrogate model is trained to learn and estimate the importance of neural design parameters at each iteration of the ENAS. The learned importance scores are used to guide the evolutionary operators in a way that leads the search algorithm to high-quality NN design spaces.

\section{Background and preliminaries}\label{sec:prelim}

\subsection{Overview on Evolutionary Search Algorithms}
\begin{figure}[h]
\centering
  \includegraphics[width=0.48\textwidth]{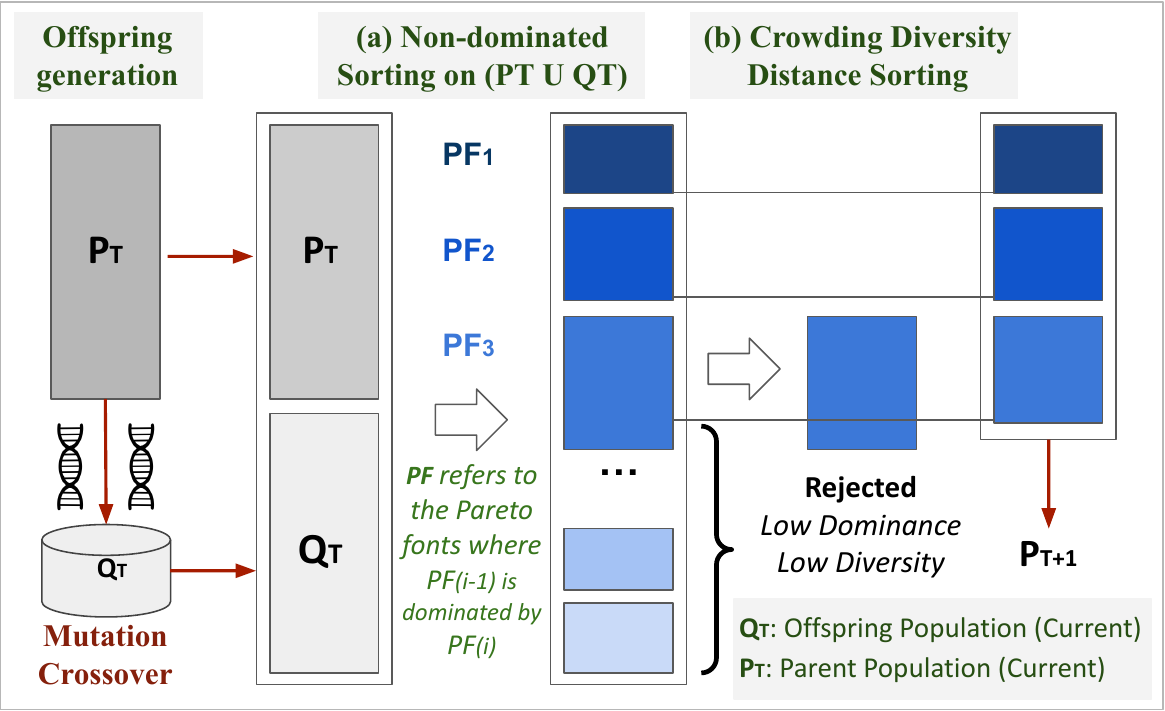}
  \caption{NSGA-II selection and evolution mechanisms. $P_T$ corresponds to current population at iteration T, $Q_T$ is the offspring population at iteration T. $PF_{(1:N)}$ are the Pareto fronts retrieved by the non-dominates sorting algorithm -- such that genomes in $PF_i$ dominate the genomes in $PF_{i+1}$.}
  \label{fig:nsga_selection}
\end{figure}

An \textit{Evolutionary Algorithm} (EA) is a metaheuristic based on the natural selection theory in biological evolution, where the best individuals survive and reproduce. This search paradigm is widely used to solve NP-hard problems to quickly retain good quality solutions while ensuring a broad exploration of gene diversity. Among the diverse EA approaches in literature \cite{vikhar2016evolutionary}, we focus on explaining the NSGA-II variant, which is the most employed one to solve NAS problems \cite{liu2021survey}. The workflow of NSGA-II is depicted in Figure \ref{fig:nsga_selection}. 

The algorithm starts first by initializing a population of NNs $(P_0)$ using random, Latin Hypercube Sampling (LSH) or data-driven initialization methods \cite{traore2023data}. Then, a fitness evaluation is performed for each genome\footnote{We use the terminology genome following the context of evolutionary optimization algorithms. For ENAS, a genome designates a NN architecture. Genes within the genomes designate the neural design parameters.} (i.e., NN architecture) in the population by computing the optimization objectives (e.g., accuracy, latency, energy). Based on the obtained values, an \textit{Elite selection} step is followed up by aggregating the values of the objectives and performing a Non-Dominated Sorting algorithm (NDS) \cite{deb2002fast} to assign \textit{Pareto dominance} scores to each genome in the population. However, focusing only on Pareto dominance to retain elite genomes is inefficient as the search algorithm will end up exploring just a small and limited region of the entire optimal Pareto front. To alleviate this issue, NSGA-II further characterizes genome diversity in the multi-objective space by computing a \textit{Crowding} distance to accommodate the selection of Pareto optimal and diverse genomes to provide the designer/end-user with a broad and better decision space. After the elite selection step, neighborhoods of new genomes are generated to characterize the next offspring population by means of evolutionary operators, notably mutation and crossover. Systematically, both operators are performed in random settings by creating perturbations on elite NN architecture designs to derive new ones. These perturbations can manifest as addition, modification, or suppression of neural operators. Subsequently generated populations $P_{(1:N)}$ will experience the same process until the expiration of the optimization budget, i.e., a fixed number of iterations, execution time, or after reaching a fitness function threshold. The maximum number of iterations is the most used criterion as an optimization budget.

\subsection{Design Parameters Importance Estimation for NAS}
Quantifying the importance of NN design parameters is crucial to pave the way for more explainability and interpretability of the type/degree of causality/correlation between NN design perturbations and the observed variations in performance/efficiency metrics. Within this scope, neural design parameters can be a priori ranked according to their importance to help the designer identify the most critical parameters for the target optimization objectives. For instance, the designer can focus on exploring and tuning subset of design parameters with a high potential to increase the hardware efficiency \cite{moussa2023hyperparameter} while maintaining the NN accuracy above a certain threshold. Alternatively, the importance of the NN design parameters can also be leveraged and exploited during or after the NAS optimization process to get insights into how and when NN architectures evolve to satisfy the optimization objectives. However, the drawn conclusions highly depend on the search space coverage and the search algorithm's effectiveness. The impact of NN design parameters on performances can be evaluated by systematically varying the values of individual design parameters while keeping others fixed and observing the resulting performance changes to analyze how NNs are sensitive to different design parameters. Sensitivity analysis aims to assess each parameter's significance by observing the degree of variability in performance when applying perturbations on each design parameter. It can be conducted via functional ANOVA (fANOVA) \cite{hutter2014efficient, watanabe2023ped}, Local Parameter Importance (LPI) \cite{biedenkapp2019cave, biedenkapp2017efficient}, or by fitting surrogate models such as Random Forest \cite{zheng2020rethinking}. Nonetheless, these techniques are only used for mono-objective settings (e.g., accuracy) and haven't been adapted or evaluated yet in the multi-contradictory objective context of HW-aware NAS. Thus, one must ask the following question: 
\begin{tcolorbox}[colback=blue!2!white,colframe=white!7!white]
\textit{How can existing parameter importance and sensitivity analysis methods be adapted to HW-aware NAS's multi-objective optimization context to understand the interplay between NN design parameters and their impact on the 'performance-efficiency' tradeoff?}
\end{tcolorbox}

\section{Problem formulation}
In the following, we detail our problem formulation. Specifically, the research problem of this paper is twofold:

\subsection{The HW-aware NAS Problem}\label{sec:nas_prob}
Let $\mathcal{M}$ be a NN architecture that comprises $n$ sequentially arranged computing blocks $\mathcal{B}^{n}$, each encompasses $d_n$ computing layer $\mathcal{L}^{d_n}$ (See Figure \ref{fig:sonata_encoding}). Typically, the type, order, and inter-dependency of $\mathcal{B}^{n}$ define the macro-architecture of the NN, whereas the internal specifications of $\mathcal{L}^{d_n}$ define the micro-architecture of the NN. To simplify the design, existing search spaces embedded into supernets \cite{caionce, wang2021alphanet, gong2021nasvit, bouzidi2023hadas, odema2023magnas, ghebriout2023harmonic}, assume a fixed macro-architecture while evolving the NN design around the variations of the $\mathcal{B}^{n}$ micro-architecture. The micro-architecture of the block defines the internal layer components. MBConv \cite{sandler2018mobilenetv2} or Attention \cite{dosovitskiy2020image}, and their respective design parameters, such as depth, kernel size, and Attention heads, are examples of micro-architecture layer components. Here, $\mathbb{M}$ designates the search space of the neural blocks micro-architecture.
\begin{align}
    \mathcal{M}(\cdot) = \mathcal{B}^{n} \circ \mathcal{B}^{n-1} \circ \mathcal{B}^{n-2} \circ \dots \circ \mathcal{B}^{2} \circ \mathcal{B}^{1} s.t. \;\;  \mathcal{M} \in \mathbb{M} \label{eqn:ss_block}\\ 
    \text{Where} \quad \mathcal{B}^{j} = \mathcal{L}^{d_j} \circ \mathcal{L}^{d_{j}-1} \circ \dots \circ \mathcal{L}^{2} \circ \mathcal{L}^{1}\label{eqn:ss_layer} 
\end{align}
A typical HW-aware NAS problem is conceptualized as a multi-objective optimization problem in which the aim is to search within the design space of $\mathbb{M}$ to retain NN architectures $\mathcal{M}^{*}$ that provide low inference error ($Err$), execution latency ($Lat$), and energy consumption ($Ergy$):
\begin{equation}
    \mathcal{M}^{*} = \argmin_{\mathcal{M} \in \mathbb{M}} [Err(\mathcal{M}), Lat(\mathcal{M}, \mathcal{H}), Ergy(\mathcal{M}, \mathcal{H})] \label{eqn:hw_nas_problem}
\end{equation}
Where $Err$, $Lat$, $Ergy$ designate the optimization objectives, and $\mathcal{H}$ is the Hardware configuration (e.g., Edge device) for deployment. Within this global optimization problem, we substitute and formulate another sub-problem for learning and estimating NN design parameters' importance given a set of NNs $\mathcal{M}$ and their respective objectives evaluations. This sub-problem is detailed in the following section.

\subsection{Design Parameter Importance Learning}\label{sec:imp_prob}
Let ($\mathbb{X}^T$, $\mathbb{Y}^T$) be a history set of sampled NN architectures ($\mathbb{X}^T$) and their evaluations on the underlying optimization objectives ($\mathbb{Y}^T$), respectively, during $T$ iterations of the ENAS. Let $\mathcal{E}$ be the encoding vector of $\mathcal{M}$ where:
\begin{align}
\mathcal{E} = [\pi_1, \pi_2, \cdots, \pi_m], \;\; s.t. \;\; \pi \in \{\mathbb{R}, \mathbb{K}, \mathbb{E}, \mathbb{W}, \mathbb{D}\}
\label{eq:sonata_vector}
\end{align}
Here $\pi_i$ defines a \textit{design parameter} from the neural block search space $\mathbb{M}$, that can encapsulate information on the input resolution ($\mathbb{R}$), kernel size ($\mathbb{K}$), channel expand ration (i.e., channel width) ($\mathbb{E}$), block width ($\mathbb{W}$), or block depth ($\mathbb{D}$). A design parameter $\pi_i$ is included in the encoding vector \textit{if on only if} it is a design variable of the search space $\mathbb{M}$. Given the multi-objective context of the HW-aware NAS, we aspire to learn the correlation between the variability of the NN design parameters and the obtained variations in Pareto optimality and diversity scores. In other words, we aim to discover the most influential design parameters on the Pareto front optimality and diversity. However, one question arises here: \textit{From which data can we learn this correlation and under which setting (offline vs. online)}? Ultimately, one straightforward way consists of performing online learning by leveraging the history of the ENAS data ($\mathbb{X}^T$, $\mathbb{Y}^T$) to train a surrogate model capable of estimating NN \textit{design parameters importance}. This model can be updated after each $\mathcal{G}$ iteration and re-trained with new sampled NN architectures. Here, $\mathcal{G}$ designates the update rate. The surrogate model for NN design importance learning is noted by $\theta_{K}^{\mathbb{M}, \mathcal{H}}$ and is detailed as follows:
\begin{align} 
\theta_{G}^{\mathbb{M}, \mathcal{H}}: \mathcal{E} \xrightarrow{}  \mathcal{S} \; , \; \mathcal{E}_i = encoding(x_i) \; | \; x_i \in \mathbb{X}^T \subset \mathbb{M} \label{eqn:importance_model} \\
\text{where} \;\; \mathcal{S}_i = Optimality(y_i)^{\beta_1} + Diversity_(y_i)^{\beta_2}\label{eqn:score_form} \\
s.t. \; y_i \in \mathbb{Y}^T \; | \; y_i = [Err(x_i), Lat(x_i, \mathcal{H}), Ergy(x_i, \mathcal{H})]
\end{align}

Here $\theta(\cdot$) defines a mapping function between the encoding vectors $\mathcal{E}_i$ of NN and a weighted sum of their \textit{Optimality} and \textit{Diversity} scores $\mathcal{S}_i$. For the sake of generality, $\beta_1$ and $\beta_1$ can be used as control knobs to favor optimality over diversity or vice-versa. In our case, as we are interested in both optimality and diversity, we set them as $\beta_1=\beta_2=0.5$.

We quantify the Pareto \textit{Optimality} using Pareto ranking scores as shown in equation (\ref{eqn:pareto_rank}). The \textit{Pareto\_rank} score is obtained from calculating the minimal Euclidean distance between each objective vector $y_i \in \mathbb{Y}^{T}$  and each optimal (i.e. non-dominated) objective vector $z_i^* \in \mathcal{PF}_{Ref}(\mathbb{Y}^{T})$ from a reference Pareto front $\mathcal{PF}_{Ref}$. We compute the reference Pareto front using the non-dominated sorting algorithm \cite{deb2002fast} on the objective vectors from $T$ previously evaluated population ($\mathbb{X}^{T}$, $\mathbb{Y}^{T}$). 
We note that $K$ in equation (\ref{eqn:pareto_rank}) refers to the number of optimization objectives. In our case, $K$=3, as we consider three objectives: The NN inference error ($Err$), Latency ($Lat$), and Energy consumption ($Ergy$).
\begin{align}
    Pareto\_rank(y_i) = \min_{z_i \in \mathcal{PF}_{Ref}} \sqrt{\sum_{k=1}^{K} (y_{i}^k - z_{i}^k)^2}\label{eqn:pareto_rank} \\
    Optimality_(y_i) = Pareto\_rank(y_i)
\end{align}

NN architectures with the lowest Pareto rank scores are the closest to the optimal reference Pareto front $\mathcal{PF}_{Ref}$, indicating better convergence. Thus, Low Pareto rank scores indicate better optimality and are favorable.

\begin{figure}[ht!]
\centering
    \includegraphics[width=.48\textwidth]{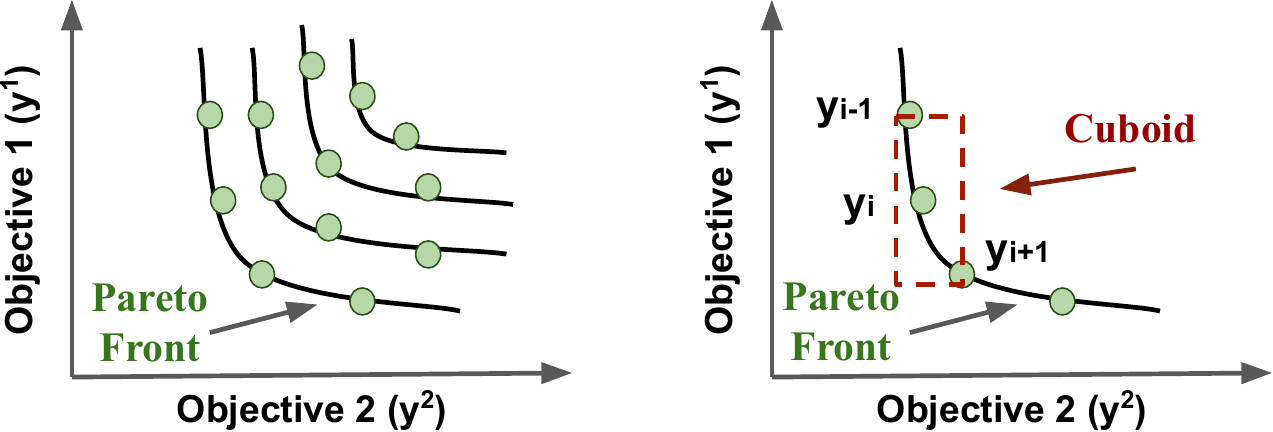}
    \caption[Computing of the crowding distance.]{Calculation of the crowding distance. To simplify the interpretability, we give an example of two optimization objectives, $y^1$ and $y^2$. The same technique can be applied to the case of more than two objectives.}
    \label{fig:crow_distance_example}
\end{figure}

We measure the Pareto \textit{Diversity} by employing the crowding distance metric introduced in NSGA-II \cite{deb2000fast} to characterize the disparity of NN architectures in the objective space (i.e., based on the disparity of their objectives vectors). The mathematical formulation of the $crow\_distance$ is detailed in equation (\ref{eqn:sonata_crowding_distance}):
\begin{align}
    crow\_distance(y_{i}) = \sum_{k=1}^{K} \frac{y_{(i+1)}^k - y_{(i-1)}^k}{y_{max}^k - y_{min}^k}\label{eqn:sonata_crowding_distance}\\
    Diversity_(y_i) = 1 - crow\_distance(y_{i})
\end{align}

\begin{figure*}[t]
\centering
    \includegraphics[width=\textwidth]{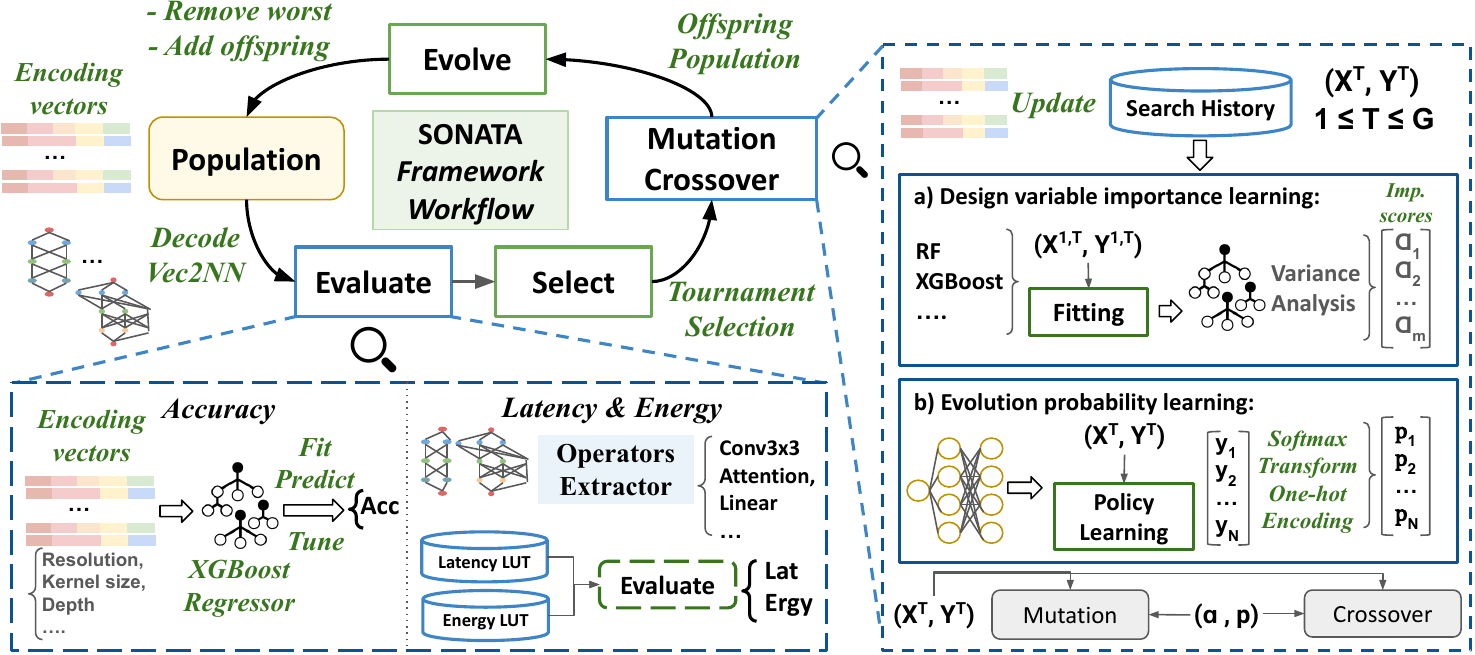}
    \caption{An overview of our novel HW-aware NAS framework: \texttt{SONATA} self-adaptive and data-driven evolutionary search process.}
    \label{fig:sonata_framework}
\end{figure*}

Where $K$ is the number of objectives, $ y_{(i+1),k}$ and $y_{(i-1),k}$ are the values of the $k$-th objective for the objective vector immediately succeeding and preceding the $i$-th objective vector, respectively. $y_{min}^k$ and $y_{max}^k$ are the minimum and maximum values of the $k$-th objective. The crowding distance of the i-$th$ objective vector $y_{i}$ expresses the average side-length of the cuboid as illustrated in Figure \ref{fig:crow_distance_example}. 

When comparing two NN architectures with different crowding distances, the NN architecture with the highest crowding distance is present in a less crowded region, thereby being more different than the others. Thus, high crowding distances indicate better diversity and are preferred.

\subsection{NN Design Parameters Importance to Guide the ENAS}
At this stage, we aim to answer the following question: \textit{How NN design parameters importance can be inferred from the mapping function $\theta(\cdot)$ between the encoding vectors the NN and the weighted sum of their optimality and diversity scores?} Particularly, we aim to infer importance scores of NN design variables $\pi$ by \textit{analyzing} the learned mapping function $\theta(\cdot)$. In other words, we aspire to measure the impact of the NN design variables $\pi$ on the performance scores by observing which design variables are critical to better fit $\theta(\cdot)$ on the search history and NN evaluations data ($\mathbb{X}^{T}, \mathbb{Y}^{T}$).

Thus, we assume that the design variables $\pi$ with the high variance in the estimations provided by $\theta(\cdot)$ are the most critical for the \textit{Optimality} and \textit{Diversity} scores of NNs.

The analysis step of the learned mapping function $\theta(\cdot)$ can be performed using methods such as variance analysis \cite{hutter2014efficient, watanabe2023ped} or by quantifying the information gain of the features coverage in tree-based ML methods \cite{chen2016xgboost, liaw2002classification}. For instance, if the mapping function $\theta(\cdot)$ is a tree-based ML model (e.g., Random Forest \cite{liaw2002classification}), the importance of design variables can be measured by the number of times each design variable has been used to split and construct nodes in the decision tree (See Figure \ref{fig:boost_tree}). Alternatively, in variance analysis with fANOVA \cite{hutter2014efficient, watanabe2023ped}, the importance of design variables can be expressed as the degree of perturbations observed in the outputs of the mapping function when varying one specific design variable while fixing the others at default values (e.g., at their minimal values).

Once the learned mapping function is analyzed, the estimated importance scores of the design parameters will be used to rank them from the least to the most important. Logically, exploring the most important NN design parameters increases the chances of retrieving NN architecture candidates with better spread and variance in the Pareto \textit{Optimality} and \textit{Diversity} scores. Accordingly, in the context of ENAS, mutation and crossover should be intensified and focused on the most critical design parameters. In a nutshell, our ultimate goal is to decide -- in a data-driven manner and under an online setting (i.e., during the search) -- which NN design parameters are worth to \textit{explore} and \textit{exploit} in the ENAS following the principal of \textit{strategic evolution from past populations by exploiting the history of genetic information}.

\section{SONATA Framework}
We propose, \texttt{SONATA}, a novel self-adaptive evolutionary framework for HW-aware Neural Architecture Search. \texttt{SONATA} leverages the search history and evaluation data from past NN populations to learn, adjust, and train ML-based models to guide the search towards Pareto optimal NN designs while sustaining minimal search overhead. The key idea of our framework is to fully reuse evaluation data generated during the search to reshape the search algorithm's understanding of how neural design variables impact the variations in Pareto optimality and diversity scores. In a data-driven fashion, \texttt{SONATA} aims to build knowledge and expertise on how the search process should evolve given the multi-objective context of HW-aware NAS. Figure \ref{fig:sonata_framework} gives an overview of our framework. We build our self-adaptive search framework upon the existing NSGA-II algorithm for multi-objective optimization by reusing fundamental components such as population initialization and tournament selection. Our contributions involve leveraging ML-based surrogate models to enhance two critical components:
\begin{itemize}
    \item (\emph{i}) \textit{Mutation and Crossover}: By using ML-based models to learn and retain the most impacting design variables that would result in diverse and optimal NNs. Moreover, we use an RL-based agent that learns -- in an unsupervised manner -- the probabilities of mutation and crossover for each genome in the selected subset $\mathbb{X}'^T$. Specifically, the role of the RL-based agent is to assign evolution probabilities to the predefined elite genomes.
    
    \item (\emph{ii}) \textit{Evaluation}: By employing inexpensive performance estimation strategies such as ML-based surrogate models to evaluate the accuracy of the sampled NN and lookup tables (LUTs) to compute the latency and energy consumption of the NN on the target hardware device. These proxy performance estimators help accelerate the fitness evaluation process -- which is typically the main bottleneck in HW-aware NAS.
\end{itemize}

Our approach contributes self-adaptive search operators upon the existing NSGA-II and can be flexible with any search space, hardware device, or target task and dataset. Furthermore, according to our preliminary analysis, the overhead of training and updating the ML-based models -- used for mutation/crossover and fitness evaluation -- is minimal and does not compromise the overall optimization time and budget. This is attributed to the fact that these ML-based models are trained on lightweight tabular data (i.e., encoding vectors $\mathcal{E}$ of NN and their respective evaluations) that don't necessitate expensive training time or computing resources.  

\begin{figure}[ht!]
\centering
    \includegraphics[width=0.48\textwidth]{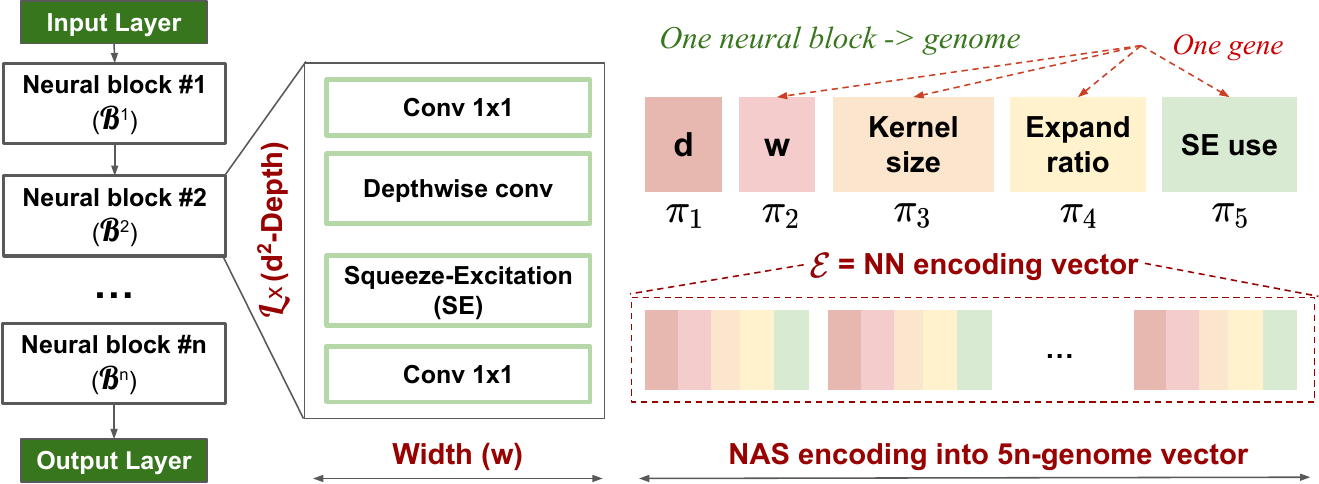}
    \caption{\texttt{SONATA} Neural network (NN) encoding scheme.}
    \label{fig:sonata_encoding}
\end{figure}
 
\subsection{Search Space Encoding and Initialization}
In ENAS, every NN architecture is represented as a \textit{genome} corresponding to an \textit{encoding vector} that embeds the neural design specifications $\mathcal{E} = [\pi_1, \pi_2, \cdots, \pi_m]$. Each genome element (gene) corresponds to one of the design variables $\pi$ of the encoding vector. We employ a direct discrete encoding as depicted in Figure \ref{fig:sonata_encoding} to represent a single NN. The same genome representation is also a feature vector to train the surrogate models in mutation/crossover and fitness evaluation steps (See Figure \ref{fig:sonata_framework}). We note that the dimension $m$ of the encoding vector depends on the search space defined in $\mathbb{M}$. In this paper, we study the case of \textit{micro-search space} $\mathcal{M}$ in which the number of neural blocks, noted $n$ in equation (\ref{eqn:ss_block}), is fixed, whereas the block components and operators in each layer {${L}^{d_j}$} of the j-$th$ block ${B}^{j}$ are searchable.

The initialization method in ENAS establishes the first design regions to explore by the search algorithm. Hence, it's extremely important to adopt an effective initialization strategy and start with diverse NN architectures such that chances of stagnation in local optima will be minimized from the beginning. For this purpose, We employ the \textit{Latin Hypercube Sampling (LHS)}, a statistical technique for efficiently sampling multi-dimensional spaces. It involves dividing each parameter's range into equally probable intervals and randomly selecting a single value from each interval for $\pi$. These values are then combined to form a sampled genome $\mathcal{E}$.

\subsection{Self-adaptive Mutation and Crossover}\label{sec:sonata_imp_model_usage}
Following the evolutionary paradigm, once a population is evaluated, a selection step follows up to render a subset of optimal genomes that will evolve to produce the next offspring population. This evolution step is ensured through mutation and crossover. Conventional mutation and crossover operators are applied to random genes of the genomes. However, this randomness often leads to a significant lack of directional focus, which becomes particularly costly when dealing with complex search spaces and expensive fitness evaluation functions, which is the case for HW-aware NAS problems. Stepping towards mitigating these challenges, we design learning-based mutation and crossover operators that target and operate on only the most important genes (i.e., design parameters). The rationale behind this approach is the following: By focusing perturbations on the most influential neural design parameters, we aim to induce high variability in the outcomes of the objectives, thereby maximizing the rates of exploration and exploitation in ENAS.

To achieve the said purpose, we leverage the search data history to learn and assign high-importance scores to neural design variables that are crucial to NN Pareto optimality and diversity. This process is illustrated in Figure \ref{fig:sonata_framework} wherein:
\begin{itemize}
    \item \textbf{Step (a)}, ML-based model is trained to learn and map importance scores to each design variable (gene).
    \item \textbf{step (b)}, RL-based policy is trained to assign mutation and crossover probabilities to each NN (genome).
\end{itemize}

Expressly, The first step addresses the '\textbf{How?}' question by identifying which design variables to evolve, whereas the second step tackles the '\textbf{Where?}' question by determining which genomes (NN architectures) are worth evolving through mutation and crossover operators.

\subsubsection{Design Parameter Importance Learning}
\begin{figure}[ht!]
\centering
    \includegraphics[width=.5\textwidth]{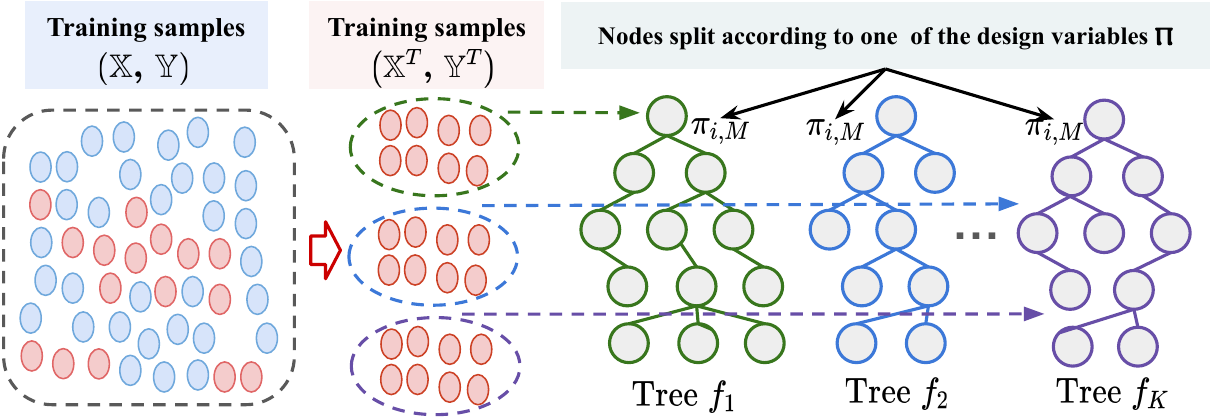}
    \caption{Nodes split mechanism in tree-based ML models (e.g. XGBoost).}
    \label{fig:boost_tree}
\end{figure}

We use the history of search data ($\mathbb{X}^T$, $\mathbb{Y}^T$) to train, tune, and update a gradient-based boosting tree \cite{chen2016xgboost} surrogate model to fit the mapping function detailed in equation (\ref{eqn:importance_model}). Each design parameter in the encoding vector $\mathcal{E}$ in equation (\ref{eq:sonata_vector}) is considered a potential feature for the node and leaf splitting during the construction of the decision trees. Specifically, we employ XGBoost to learn and construct the tree-based model $\theta(\cdot)$. The learning process follows a gradient-descent optimization where the loss function to minimize is given as follows:
\begin{align}
    \mathcal{L}oss_{\theta}(\mathcal{S}, \hat{\mathcal{S}}) = \sqrt{\frac{1}{N} \sum_{i=1}^{N} (\mathcal{S}_i - \hat{\mathcal{S}}_i))^2} + \sum_{q=1}^{Q} \mathcal{L}_2(f_q)\label{eq:xgb} \\
    \text{where} \;\; \Omega(f_q) = \gamma R + \frac{1}{2} \lambda \| \mathbf{w} \|^2 
\end{align}
Where $N$ is the number of ground-truth labels and $Q$ is the maximum number of decision trees, $\Omega$ is the $L_2$ regularization term used to control over-fitting by penalizing the complexity of the model. Specifically, $\Omega$ is defined for each tree $f_q$ where $R$ is the number of leaves, $w$ is the vector of scores in the leaf nodes, $\gamma$ and $\lambda$ are regularization parameters. Each node or leaf is associated with a unique feature, i.e., decision variable $\pi_{i} \in \mathcal{E}$. Once the learning process completes its procedure, a vector of importance scores $\alpha=[\alpha_1, \alpha_1, \dots, \alpha_m]$ is generated. Each $\alpha_i$ designates an importance score of its respective design variable $\pi_{i} \in \mathcal{E}$. The importance score is computed based on the number of times a leaf or node, in decision trees, was split based on $\pi_i$ as illustrated in Figure \ref{fig:boost_tree}.

\subsubsection{Evolution Probability Learning}
Learning the importance of the NN design parameter helps pinpoint which genes are worth exploring during the ENAS process. However, there remains another source of uncertainty that comes from the probability rate at which evolution operators should be applied. Dynamic probability sampling for mutation and crossover has been shown promising for ENAS \cite{xue2021self}. 

In a similar manner, we replace the typical static evolution probabilities with a learning-based probability sampling method. Specifically, we train a \textit{Policy agent} (See Figure \ref{fig:sonata_framework}) to learn the probabilities of mutation/crossover per genome (i.e., NN architecture). We formulate the evolution probability learning as a sequential decision-making process, where an agent, a neural network denoted as $\psi$, makes a sequence of decisions about selecting appropriate evolution probabilities $p=[p_1, p_2,\dots,p_N]$ according to the gain in Pareto optimality and diversity scores in $\mathcal{S}$. We design the network agent $\psi$ with two fully connected (FC) layers. The agent $\psi$ takes as inputs the encoding vectors ($\pi^*$) of the predefined elite genomes issued from the tournament selection step. 

The two FC layers perform feature extraction from the NN encoding vectors to select appropriate evolution probabilities for each genome. To enhance the policy learning and decision-making for evolution probability sampling, we train our agent network $\psi$ in a Reinforcement Learning fashion \cite{zoph2018learning} using a reward function $\mathcal{R}_{\psi}$, detailed as follows:
\begin{equation}
    \mathcal{R}_{\psi} = \frac{1}{N} \sum_{i=1}^{N} \left( -\left(\mathcal{S}_i - \text{max}(\mathcal{S}_{(1,\dots,N}) \right) \times \log(\text{soft\_prob}_i) \right)
\end{equation}
where $\mathcal{S}$ denotes the set of the Pareto optimality and diversity scores, $sof\_prob$ are the Softmax probabilities predicted by the agent network $\psi$ for the given inputs of encoding vectors. We employ the policy gradient algorithm of the N2N paradigm proposed in \cite{ashok2018nn} to optimize our agent network $\psi$.

\subsection{Surrogate-assisted Fitness Evaluation}\label{sec:sonata_fitness_eval}
To accelerate the fitness evaluation, we leverage surrogate models based on XGBoost \cite{chen2016xgboost} for regression problems. Prior works in surrogate modeling, such as NAS-Bench-301 \cite{siems2020bench} and \cite{bouzidi2022performance}, have proven the generalization and rank-preserving power of XGBoost. In our framework \texttt{SONATA}, we employ XGBoost to model the accuracy of NNs. Prior to the search process, we randomly sampled and evaluated 500 NNs on the validation dataset to get their ground-truth accuracy measurements. Then, we use the sampled NNs and their accuracy measurements to tune and train the XGBoost model. We note that the encoding vectors of the sampled NNs are used as feature vectors for the XGBoost model.

For hardware-related metrics, we use lookup tables (LUTs) by inspecting and analyzing the commonly used operators (e.g., convolution, attention, fully connected) across the studied search spaces in this work. We then deploy and profile the operators on the target hardware devices to get latency and energy consumption measurements. The reason for employing LUTs is that recent and effective search spaces in \cite{caionce, cai2018proxylessnas, wang2021alphanet, gong2021nasvit} share common and similar neural operators and configurations. Thus, retrieving these common operations and measuring their hardware overheads once and for all is more convenient to reuse them across diverse search spaces.

\section{Evaluation}
In this section, we conduct experiments to showcase the merit of \texttt{SONATA} to render optimal NN architectures under a low optimization budget. We evaluate the generality of our framework across diverse NAS search spaces and Edge hardware devices. Specifically, we experiment on SOTA CNN and Transformer search spaces for image classification on the large ImageNet-1k dataset on three different edge GPUs from NVIDIA. Following, we provide more details on the experiment setup and evaluation results.

\subsection{Experimental Setup}
\noindent{\large \textcircled{\small 1}} \textbf{NAS Search Spaces.}
We study the case of four SOTA search spaces designed initially for resource-constrained computing systems. Specifically, we built our NAS search spaces upon the baselines introduced in AlphaNet \cite{wang2021alphanet}, Once-for-all (OFA) \cite{cai2019once}, ProxylessNAS \cite{cai2018proxylessnas}, and NASViT \cite{gong2021nasvit}. The three first ones are based on CNN architectures. Notably, OFA and ProxylessNAS are built upon MobileNet-V3 \cite{mobilenetv3}, AlphaNet upon FBNet \cite{wu2019fbnet}, while NASViT is a CNN-Transformer hybrid architecture combining MBConv blocks from \cite{mobilenetv3} and Transformer blocks from \cite{dosovitskiy2020image}. \\

\noindent{\large \textcircled{\small 2}} \textbf{Hardware Devices Settings.}
We validate our method on three edge devices from NVIDIA:
\begin{enumerate}
    \item \textsc{Jetson AGX Xavier} equipped with an NVIDIA Carmel Arm-64bit CPU and a Volta GPU of 512 GPU cores.
    \item \textsc{Jetson TX2} composed of an NVIDIA Denver 64Bit and ARM-A57 CPU cores and a Pascal GPU with 256 GPU cores.
    \item \textsc{Jetson Nano} that comprises an ARM-A57 CPU and a Maxwell GPU of 128 GPU cores.
\end{enumerate} 

NNs have been implemented and deployed using TensorRT 8.4 as a high-performance SDK from NVIDIA \cite{vanholder2016efficient}. We map the NN entirely on the GPU as the target computing unit under FP32 data precision and \texttt{MAXN} \texttt{DVFS} setting. \\

\noindent{\large \textcircled{\small 3}} \textbf{Evolutionary Search Settings.}
We build \texttt{SONATA} on top of the existing NSGA-II algorithm \cite{deb2002fast}. We keep the main algorithmic components. The changes we made are at the levels of evolution operators and fitness evaluation as discussed in Sections \ref{sec:sonata_imp_model_usage} and \ref{sec:sonata_fitness_eval}. We fix the population size to 300 and run the optimization framework for $G=10$ generations. After every two generations, we retrain and update the surrogate models and the network agent for the evolution operators. Each NAS search space contains more than 30 design parameters. Among these neural design parameters, we focus on the top five (05) with the highest importance scores. The network agent $\psi$ selects the evolution probabilities within a predefined $[0.3, 1.0]$ range. 

We use the native NSGA-II --without our modifications-- as a baseline search algorithm for comparison in the following sections. To ensure a fair comparison, we use the same population initialization method (i.e., LSH), fitness evaluation strategy, elite selection, and optimization budget (i.e., population size and number of generations). We also tuned the mutation and crossover probabilities and found a combination of 0.5-0.5 as an optimal setting for this baseline.

\begin{figure}[ht]
\centering
    \includegraphics[width=.4\textwidth]{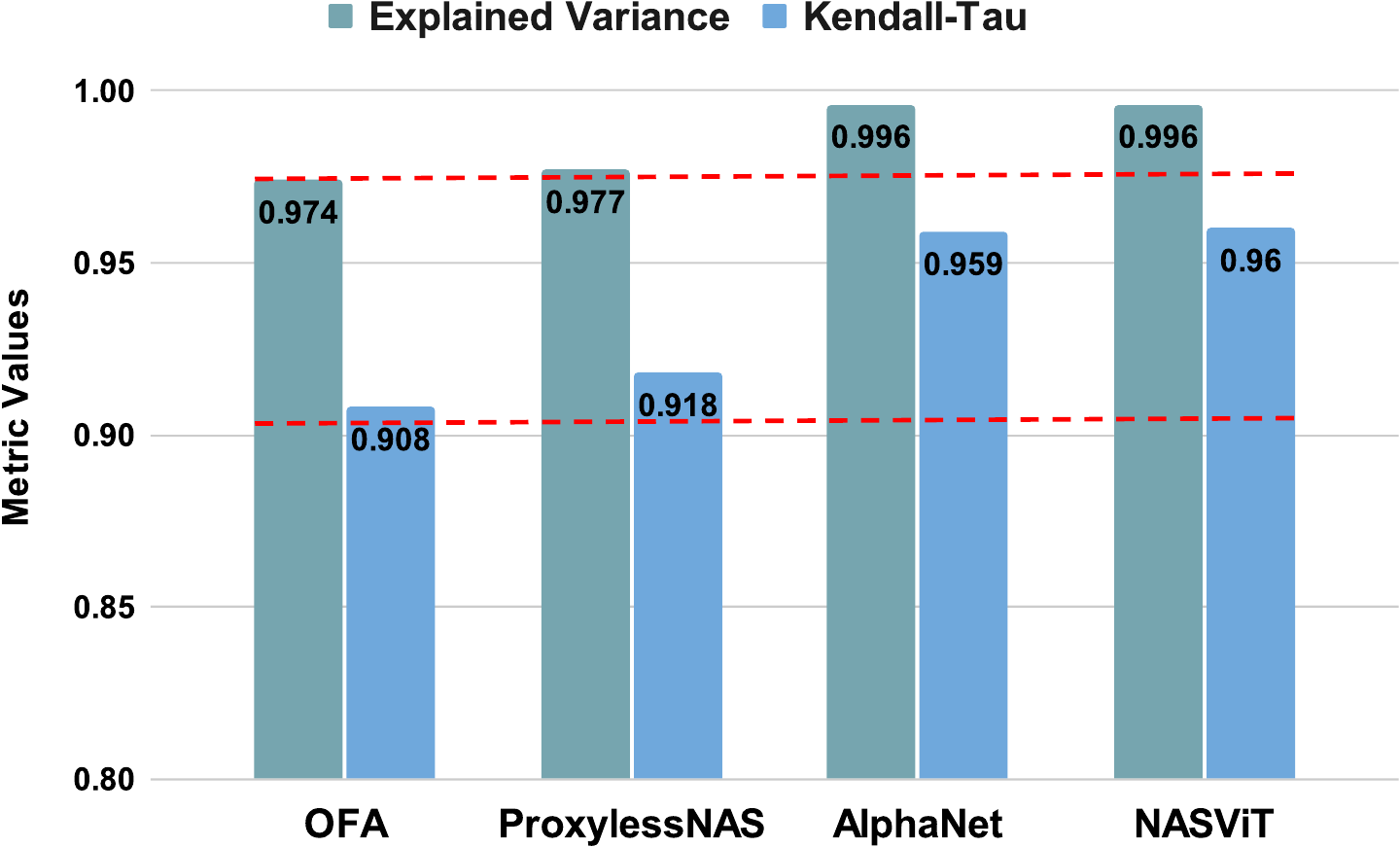}
    \caption{Performances of the XGBoost-based surrogate models used to estimate the accuracy of the sampled NNs in the fitness evaluation step of \texttt{SONATA}. We report the explained variance and Kendall-Tau rank correlation coefficients -- High values indicate accurate estimations of NN accuracy.}
    \label{fig:sonata_accuracy_model}
\end{figure}

\subsection{Surrogate models analysis} 
In this section, we discuss the effectiveness of the surrogate models employed in our framework, notably:
\begin{itemize}
    \item (\emph{i}) The surrogate model used for NN accuracy estimation in the fitness evaluation step of \texttt{SONATA}.
    \item (\emph{ii}) The surrogate model used for design variable importance learning and estimation in the Mutation/Crossover evolution step of \texttt{SONATA}.
\end{itemize}

Firstly, we discuss the performance of the XGBoost surrogate model used to estimate the accuracy of the sampled NN during the search. As shown in Figure \ref{fig:sonata_accuracy_model}, XGBoost gives relatively accurate estimations in all NAS search space settings. The explained variances, also known as R2, and Kendall-Tau rank correlation coefficient are above 0.80. This result indicates a strong prediction accuracy and model rank preservation. Our preliminary results also have shown a prediction accuracy higher than 95\% in all the studied search spaces, further stipulating the generalization and rank-preserving power of XGBoost \cite{chen2016xgboost}. We also note that the shared macro-architecture structure in SOTA NAS search spaces highly contributes to the performance of the surrogate models used for accuracy \cite{siems2020bench}. For unstructured search spaces with uneven NN macro- and micro-architectures, fitting a relatively simple surrogate model to predict the NN accuracy can be challenging. Thus, employing a strong predictor like XGBoost is highly recommended.

\begin{figure}[ht!]
\centering
    \includegraphics[width=.5\textwidth]{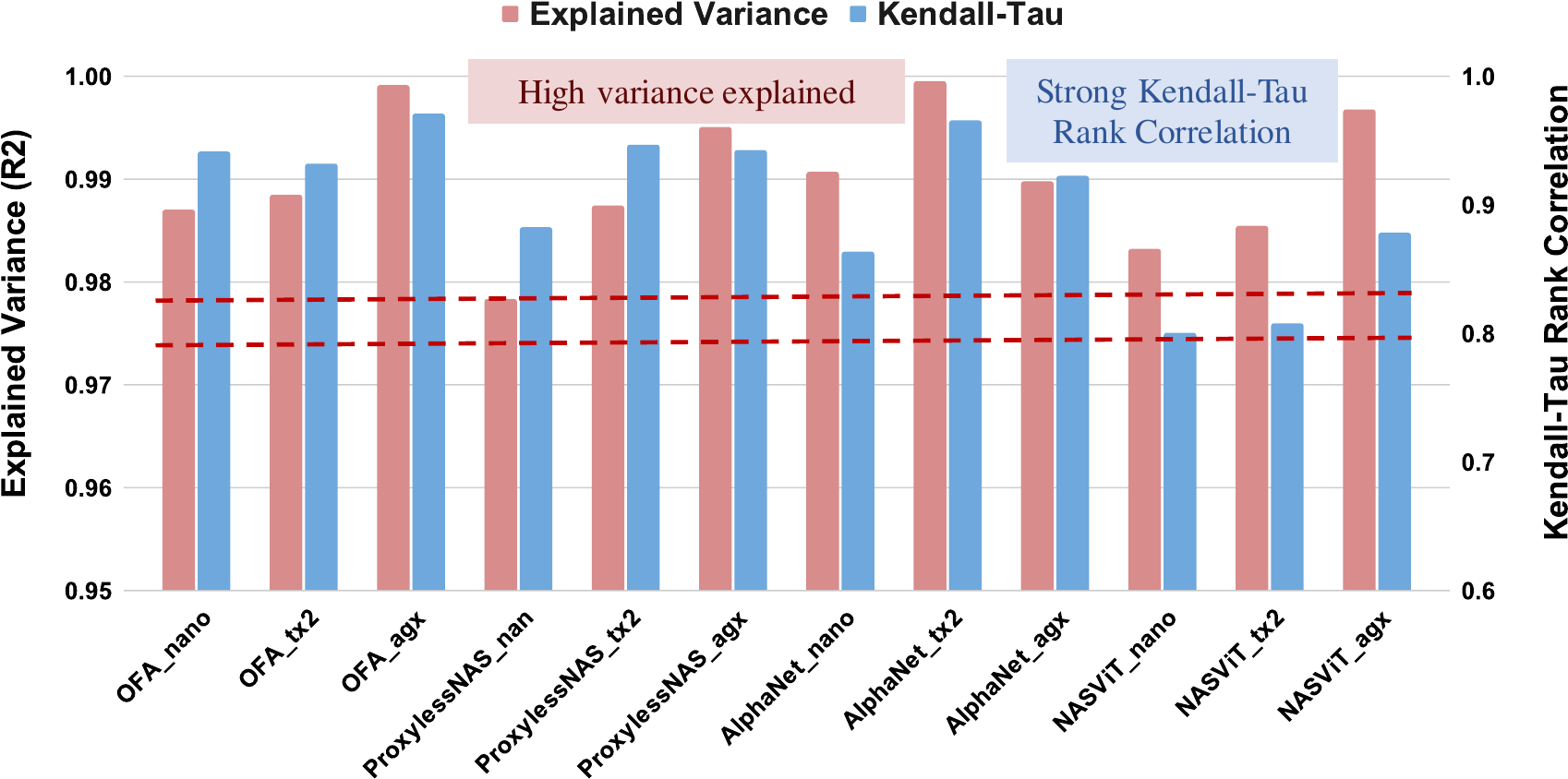}
    \caption{Performances of the surrogate models used to learn the importance of NN design parameters in the mutation/crossover step of \texttt{SONATA}. We report the explained variance and Kendall-Tau rank correlation coefficients -- High values indicate accurate estimations of NN design parameter importance.}
    \label{fig:sonata_perf_imp_models}
\end{figure}

\begin{table*}[tp]
\centering
\caption{Comparison between the optimization efficiency (i.e., in terms of convergence and diversity of the approximated Pareto front) of the baseline static ENAS NSGA-II \cite{deb2002fast} and our self-adaptive \texttt{SONATA}. We assess convergence and diversity using three commonly used metrics in multi-objective optimization contexts: Hypervolume, IGD distance, and dominance ratio.}
\label{tab:sonata_optim_metrics}
\scalebox{1.2}{
\begin{tabular}{cl|cccccc} 
\hline
\multirow{2}{*}{\begin{tabular}[c]{@{}c@{}}\textbf{HW}\\\textbf{Device}\end{tabular}} & \multicolumn{1}{c|}{\multirow{2}{*}{\begin{tabular}[c]{@{}c@{}}\textbf{NAS Search }\\\textbf{Spaces}\end{tabular}}} & \multicolumn{3}{c}{\textbf{Static ENAS NSGA-II \cite{deb2002fast} ~}} & \multicolumn{3}{c}{\textbf{Our Adaptive \texttt{SONATA} }} \\ 
\cline{3-8}
 & \multicolumn{1}{c|}{} & \textbf{Hypervolume} & \textbf{IGD} & \begin{tabular}[c]{@{}c@{}}\textbf{Dominance }\\\textbf{Ratio}\end{tabular} & \textbf{Hypervolume} & \textbf{IGD} & \begin{tabular}[c]{@{}c@{}}\textbf{Dominance }\\\textbf{Ratio}\end{tabular} \\ 
\hline
\multirow{4}{*}{\begin{tabular}[c]{@{}c@{}}Jetson \\Nano\end{tabular}} & OFA \cite{caionce} & 1725509 & 1.484 & 0.327 & \textbf{1733316} & \textbf{0.1305} & \textbf{0.826} \\
 & ProxylessNAS \cite{cai2018proxylessnas} & 1464862 & 1.072 & 0.071 & \textbf{1469478} & \textbf{0.0233} & \textbf{1.0} \\
 & AlphaNet \cite{wang2021alphanet} & \textbf{2336429} & 1.592 & \textbf{0.687} & 2330159 & \textbf{0.8175} & 0.305 \\
 & NASViT \cite{gong2021nasvit} & 1547510 & 3.040 & 0.447 & \textbf{1548843} & \textbf{0.4109} & \textbf{0.853} \\ 
\hline
\multirow{4}{*}{\begin{tabular}[c]{@{}c@{}}Jetson \\TX2\end{tabular}} & OFA \cite{caionce} & 1756847 & 2.299 & 0.399 & \textbf{1757666} & \textbf{0.2964} & \textbf{0.866} \\
 & ProxylessNAS \cite{cai2018proxylessnas} & 1485138 & 2.291 & 0.170 & \textbf{1495840} & \textbf{0.1373} & \textbf{0.936} \\
 & AlphaNet \cite{wang2021alphanet} & \textbf{2479025} & 0.978 & \textbf{0.806} & 2476035 & \textbf{0.3215} & 0.372 \\
 & NASViT \cite{gong2021nasvit} & 1809648 & 6.435 & 0.417 & \textbf{1817793} & \textbf{0.5333} & \textbf{0.809} \\ 
\hline
\multirow{4}{*}{\begin{tabular}[c]{@{}c@{}}Jetson \\AGX\end{tabular}} & OFA \cite{caionce} & 1842773 & 0.282 & 0.415 & \textbf{1845282} & \textbf{0.0965} & \textbf{0.980} \\
 & ProxylessNAS \cite{cai2018proxylessnas} & 1569035 & 0.850 & 0.043 & \textbf{1572546} & \textbf{0.0027} & \textbf{1.0} \\
 & AlphaNet \cite{wang2021alphanet} & \textbf{2558571} & 0.658 & \textbf{0.826} & 2561133 & \textbf{0.2776} & 0.579 \\
 & NASViT \cite{gong2021nasvit} & 2268507 & 3.854 & 0.524 & \textbf{2270813} & \textbf{0.8185} & \textbf{0.709} \\
\hline
\end{tabular}
}
\end{table*}

\begin{figure*}[ht]
\centering
    \includegraphics[width=1.0\textwidth]{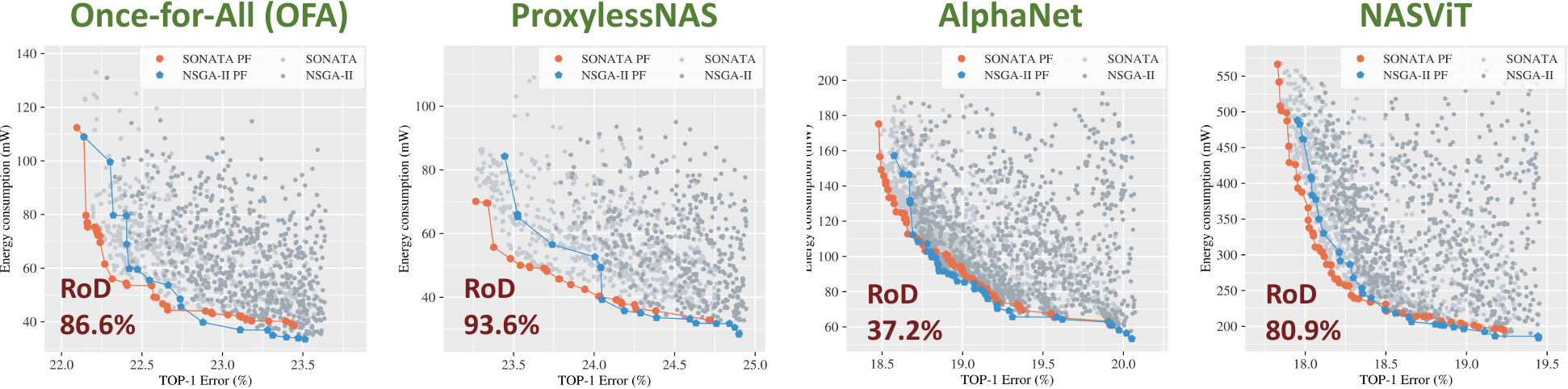}
    \caption[Comparing the optimization results of \texttt{SONATA} Vs. NSGA-II.]{Comparing the optimization results of \texttt{SONATA} Vs. NSGA-II. The results are reported on all of the studied the search spaces on the NVIDIA Jetson TX2. The TOP-1 error are reported for the ImageNet-1k dataset. RoD denoted the Ratio of dominance metric from Table \ref{tab:sonata_optim_metrics}.}
    \label{fig:sonata_paretos}
\end{figure*}

Secondly, in Figure \ref{fig:sonata_perf_imp_models}, we report the performances of the surrogate models $\theta(\cdot)$ used for NN design parameters importance learning (Recall Section \ref{sec:imp_prob}). $\theta(\cdot)$ progressively learns to estimate the importance of NN design variables to guide the mutation/crossover in \textbf{Step (a)} of the evolution (As depicted in Figure \ref{fig:sonata_framework}). We note that these surrogate models also need to leverage information on NN accuracy and HW efficiency metrics to compute Pareto optimality and diversity scores as previously explained in Section \ref{sec:sonata_imp_model_usage}. These scores highly depend on the NAS search space and the target HW device and represent the ground-truth labels for $\theta(\cdot)$. The pairs (NAS search space $\times$ HW device) are indicated in the x-axis of the graph depicted in Figure \ref{fig:sonata_perf_imp_models}.

As shown in Figure \ref{fig:sonata_perf_imp_models}, the high values of the explained variances and Kendall-Tau rank correlation coefficients indicate strong performances and high fitness of $\theta(\cdot)$ models. The Kendall-Tau correlation coefficients are slightly lower than those in Figure\ref{fig:sonata_accuracy_model}. The reason behind this observation is the complexity of the correlation in the scores used to train $\theta(\cdot)$. This complexity arises from the contradictory nature of the involved objectives, notably NN accuracy and HW efficiency. According to our preliminary analysis, we notice the existence of a weak correlation between accuracy and hardware-related metrics (i.e., latency and energy). Furthermore, we noticed that the diversity term in (\ref{eqn:score_form}) added another layer of complexity. Nevertheless, we noticed that the results of \texttt{SONATA} can be less diverse without including the diversity term in (\ref{eqn:score_form}) -- thus, we opted to keep it along the optimality term to obtain an optimal balance of both. 

\subsection{\texttt{SONATA} Optimization Efficiency} 
In multi-objective optimization problems, both convergence and diversity are of high priority and importance. To showcase the merit of our framework \texttt{SONATA} compared to baseline ENAS algorithms, such as NSGA-II, we employ different performance evaluation metrics widely used to assess optimality and diversity:
\begin{itemize}
    \item (\emph{i}) \textit{Hypervolume} that measures the portion of the objective space dominated by the Pareto front obtained by the search algorithm. 
    \item (\emph{ii}) \textit{Inverted Generational Distance (IGD)} that measures the average distance from each point in the optimal Pareto front to its nearest point in the Pareto front obtained by the search algorithm. As the optimal Pareto front is unknown for the NP-hard problems, we approximate it by merging the Pareto fronts of our \texttt{SONATA} and the baseline NSGA-II.
    \item (\emph{ii}) \textit{Dominance ratio} that computes the ratio of non-dominated solutions from the obtained Pareto front by the search algorithm contributing to the optimal Pareto front -- approximated as previously mentioned.
\end{itemize}

We summarize the results of this evaluation in Table \ref{tab:sonata_optim_metrics}. We note that high hypervolume and dominance ratio values indicate an optimal balance between convergence and diversity, whereas low IGD values stipulate optimal convergence. As reported in Table \ref{tab:sonata_optim_metrics}, our \texttt{SONATA} generally outperforms the baseline search algorithm NSGA-II in all optimization metrics. This is attributed to the ability of \texttt{SONATA} to focus on the most rewarding NN design variables while expanding the scope of exploitation and exploration by dynamically varying the mutation/crossover probabilities via the RL-based agent network $\psi$. 

More interestingly, the baseline NSGA-II only outperforms \texttt{SONATA} in the AlphaNet \cite{wang2021alphanet} search space. However, by observing the exploration results of Figure \ref{fig:sonata_paretos}, we can see that \texttt{SONATA} provides more accurate NNs with the same HW execution cost as NSGA-II. Thus, we highlight the importance of visual analysis when assessing the performance of optimization algorithms for HW-aware NAS. Quantitative and qualitative analysis must be jointly used to determine the effectiveness of search algorithms in HW-aware NAS. For instance, in the AlphaNet case, if NN accuracy is of the highest priority among other objectives, the designer/user may be more interested in the results of \texttt{SONATA}, as they provide more accurate NNs despite failing to outperform the baseline NSGA-II in the quantitative analysis of Table \ref{tab:sonata_optim_metrics}. 

From the exploration results in Figure \ref{fig:sonata_paretos}, we can also see how \texttt{SONATA} was able to explore diverse accuracy-efficiency trade-offs. Notably, by intensifying the search in the region of interest -- the extreme left depicts the highest accuracy and lowest energy consumption. We can observe that \texttt{SONATA} optimization reaches NN designs that provide better accuracy and low energy consumption than those reached by the baseline NSGA-II. For instance, in the ProxylessNAS case, the Pareto front of \texttt{SONATA} provides an accuracy improvement by \textbf{$\sim$0.25\%} and energy gains of \textbf{$\sim$2.42x}. These observations further demonstrate the merit of adapting the evolution operators to intensify the search around the most important and promising NN design parameters.

\section{Conclusion and Future Work}
In this paper, we have explored the prospect of leveraging ML-based methods to enhance the efficiency of evolutionary NAS frameworks. Specifically, we built our motivation upon the assumption that not all NN design variables can substantially improve performance and hardware efficiency. We thus argue that ENAS search algorithms can focus on exploring the most rewarding design variables to accelerate the convergence and discovery of optimal NN design with better performance-efficiency tradeoffs. 

Following this assumption, we proposed \texttt{SONATA}, a self-adaptive evolutionary framework for multi-objective HW-aware NAS. Our framework aims to reduce the randomness and uncertainty of conventional evolutionary operators (i.e., mutation and crossover) and replace them with data-driven ML-based methods. As such, we use the search data history and tree-based ML methods to progressively learn the importance of NN design variables pursuing optimality and diversity in the multi-objective context of HW-aware NAS. The learned importance scores are then used to select a subset of the most critical NN design variables on which mutation and crossover should be applied. Furthermore, we employ an RL-based agent that assigns dynamic mutation and crossover probabilities to evolve only the most promising NN architectures. We also employ ML-based and LUT as surrogate models to accelerate the fitness evaluation step.

Evaluation results have shown the merit of our approach with an accuracy improvement up to \textbf{$\sim$0.25\%} and latency/energy gains up to \textbf{$\sim$2.42x}. \texttt{SONATA} has also seen up to \textbf{$\sim$93.6\%} Pareto dominance over the native NSGA-II. 

We have made a first step towards a self-adaptive ENAS by leveraging ML-based methods to enhance evolutionary operators and fitness evaluation. However, data-driven ML-based methods can be extended to enhance other NAS components, notably:
\begin{itemize}
    \item Search space by adding, altering, or removing NN design parameters based on their importance scores to the performances, prior \cite{biedenkapp2019cave} or during the search (our work).
    \item Search strategy by leveraging the dynamic reconfigurability of the search algorithm hyperparameters \cite{lange2023discovering}.
    \item Fitness evaluation by training surrogate predictors to estimate NN performances under offline or online settings.
\end{itemize}
 
Future works can extend our \texttt{SONATA} by enhancing the knowledge of the importance of NN design variables with GNN or Transformers as introduced in \cite{lange2023discovering}. Incorporating preference-based evolutionary operators to favor specific objectives and intensify the search around certain parameter values without imposing strict constraints or pruning the search spaces a priori or progressively is an important step for more self-adaptive HW-aware ENAS frameworks. 

\texttt{SONATA} can be used offline to adjust the search space before the search. It can also be used in an online setting, as we proposed in this paper, or by hierarchizing the exploration process for multi-level search spaces. It will be possible to co-optimize the NN and the HW by simultaneously setting their corresponding design parameters.

\bibliographystyle{IEEEtran}
\bibliography{references}

\end{document}